\documentclass[letterpaper]{article} 
\usepackage{aaai25} 
\usepackage{times}  
\usepackage{helvet}  
\usepackage{courier}  
\usepackage[hyphens]{url}  
\usepackage{graphicx} 
\usepackage{natbib}  
\usepackage{caption} 
\usepackage{algorithm}
\usepackage{algorithmic}

\usepackage{newfloat}
\usepackage{listings}

\usepackage{booktabs}
\usepackage{tcolorbox}
\tcbuselibrary{breakable} 
\tcbuselibrary{skins} 
\usepackage{colortbl}
\usepackage{xcolor}
\usepackage{lipsum}
\usepackage{tabularx} 
\usepackage{multirow} 
\usepackage{array}    
\usepackage{enumitem}
\usepackage{makecell}
\usepackage{subcaption}
\usepackage{amsmath}
\usepackage{textcomp}
\usepackage{amssymb}

\usepackage{xcolor}

\DeclareCaptionStyle{ruled}{labelfont=normalfont,labelsep=colon,strut=off} 
\floatstyle{ruled}
\newfloat{listing}{tb}{lst}{}
\floatname{listing}{Listing}
\title{SS-GEN: A Social Story Generation Framework with Large Language Models}

\author {
    Yi Feng\textsuperscript{\rm 1},
    Mingyang Song\textsuperscript{\rm 2},
    Jiaqi Wang\textsuperscript{\rm 3}\footnotemark[1],
    Zhuang Chen\textsuperscript{\rm 4},
    Guanqun Bi\textsuperscript{\rm 5} \\
    Minlie Huang\textsuperscript{\rm 5},
    Liping Jing\textsuperscript{\rm 1}\thanks{Corresponding Author},
    Jian Yu\textsuperscript{\rm 1}
}
\affiliations {
    \textsuperscript{\rm 1}Beijing Key Laboratory of Traffic Data Mining and Embodied Intelligence, Beijing Jiaotong University\\
    \textsuperscript{\rm 2}Tencent Hunyuan, Beijing\\
    \textsuperscript{\rm 3}Jarvis Research Center, Tencent YouTu Lab\\
    \textsuperscript{\rm 4} School of Computer Science and Engineering, Central South University\\
    \textsuperscript{\rm 5}CoAI Group, DCST, IAI, BNRIST, Tsinghua University\\
    yifeng@bjtu.edu.cn
}

\usepackage{bibentry}

\begin{document}

\maketitle

\begin{abstract}
Children with Autism Spectrum Disorder (ASD) often misunderstand social situations and struggle to participate in daily routines. Social Stories™ are traditionally crafted by psychology experts under strict constraints to address these challenges but are costly and limited in diversity. As Large Language Models (LLMs) advance, there's an opportunity to develop more automated, affordable, and accessible methods to generate Social Stories in real-time with broad coverage. However, adapting LLMs to meet the unique and strict constraints of Social Stories is a challenging issue. To this end, we propose \textbf{SS-GEN}, a \textbf{S}ocial \textbf{S}tory \textbf{GEN}eration framework with LLMs. Firstly, we develop a constraint-driven sophisticated strategy named \textbf{\textsc{StarSow}} to hierarchically prompt LLMs to generate Social Stories at scale, followed by rigorous human filtering to build a high-quality dataset. Additionally, we introduce \textbf{quality assessment criteria} to evaluate the effectiveness of these generated stories. Considering that powerful closed-source large models require very complex instructions and expensive API fees, we finally fine-tune smaller language models with our curated high-quality dataset, achieving comparable results at lower costs and with simpler instruction and deployment. This work marks a significant step in leveraging AI to personalize Social Stories cost-effectively for autistic children at scale, which we hope can encourage future research on special groups.

\end{abstract}

%

\section{Introduction}

\begin{figure}[t]
  
  \centering
  \includegraphics[width=\columnwidth]{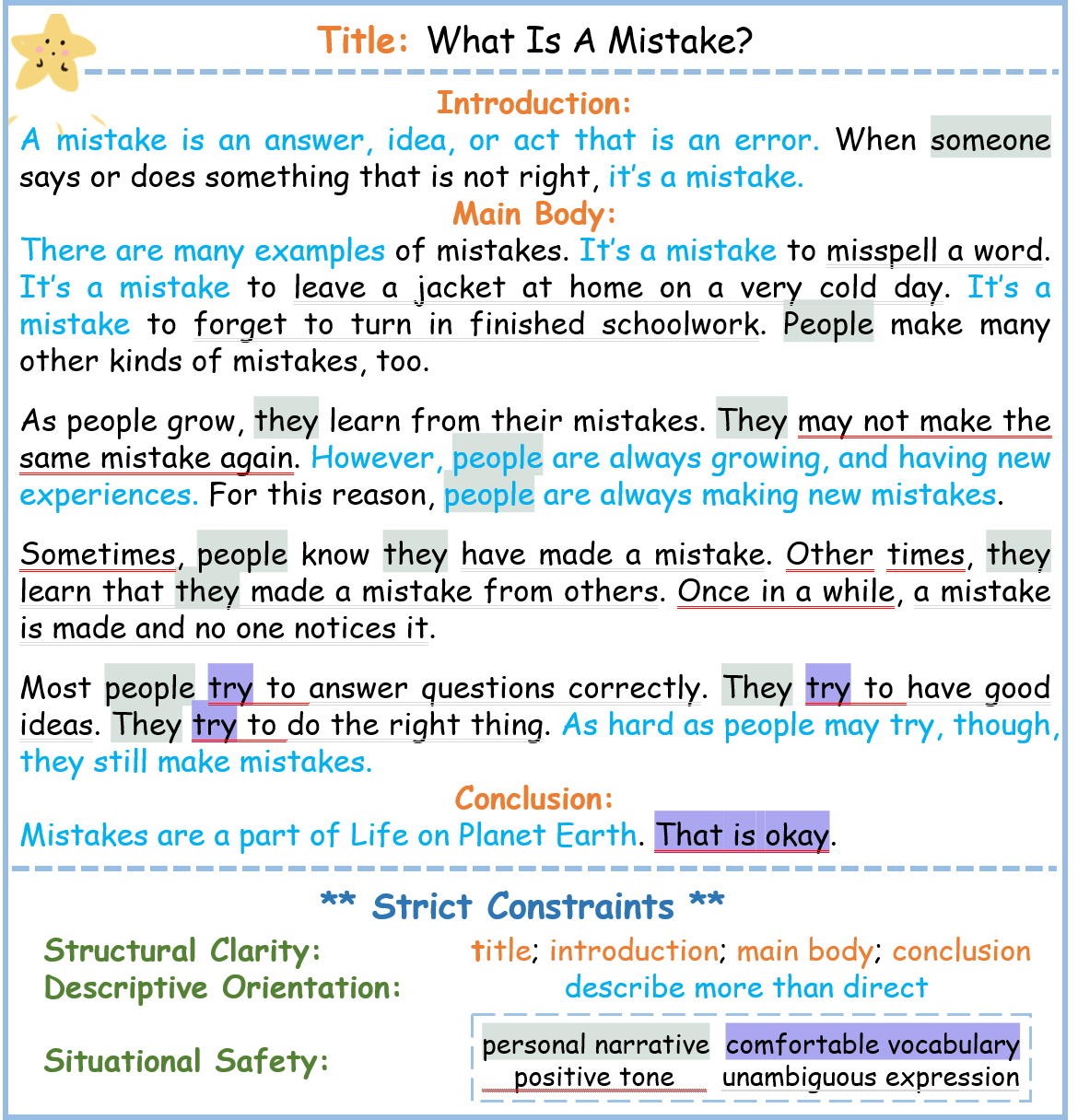} 
  \caption{A Social Story excerpted from {\it The New Social Story Book} written by Carol Gray \citep{gray2010new}, which adhere to strict constraints that we summarized based on professional guidelines: Structural Clarity, Descriptive Orientation, Situational Safety.}
  \label{fig:intro-SS}
\end{figure}

Social Story™\footnote{As the social-skill intervention tool, Social Stories™ are widely used to specifically help individuals with ASD understand and navigate specific social situations. When capitalized, a Social Story refers to a Social Story™ following the criteria in 10.2 Tutorial \citep{gray2010new}.}, pioneered by \citet{gray1993social}, helps children with autism\footnote{Autism Spectrum Disorder (ASD) is a lifelong neurodevelopmental disorder characterized by impairments in social interaction, communication, and restricted or repetitive behaviors, affecting millions of children (approximately 1/100) worldwide with the number continuously growing \citep{zeidan2022global}.} define the social context of anticipated or experienced situations, putting them into perspective and then developing statements on how they feel and what actions to take in response, thereby positively improving their social understanding and behavioral outcomes. For example, the story in Figure~\ref{fig:intro-SS} helps autistic children understand and manage the stress and anxiety triggered by mistakes.

These stories are traditionally created by psychologists following the 10.2 tutorial \cite{gray2010new}, which we summarize as three strict constraints: \textbf{Structural Clarity} to organize the title, introduction, main body and conclusion for clear social cues; \textbf{Descriptive Orientation} to describe more than direct, being patient and humility; \textbf{Situational Safety} to emphasize an appropriate perspective, positive tone, unambiguous language and precise vocabulary, ensuring the reassuring quality and safeguarding the self-esteem of autistic youth. These handcrafted stories serve as a library for caregivers to customize according to the specific needs of ASD children, but are costly and limited in diversity and scope.

With rapid advancements of Large Language Models (LLMs), it's promising to automatically generate Social Stories with LLMs. Existing research has explored generating suspenseful \citep{xie2024creating}, playable \citep{nasir2024word2world}, age-appropriate \citep{valentini2023automatic, bhandari2023trustworthiness} or legal-concept informative \citep{jiang2024leveraging} stories, primarily for adults or general children. However, adapting LLMs to meet the unique and strict constraints of Social Stories is extremely challenging and rarely a topic of interest, despite the importance of this early intervention tool for improving social skills and enhancing the life quality of such special group \citep{koegel2014importance}.

In this work, we introduce \textbf{SS-GEN}, a \textbf{S}ocial \textbf{S}tory \textbf{GEN}eration framework based on their unique constraints. (1) Initially, we develop a constraint-driven strategy named \textbf{\textsc{StarSow}} (Figure~\ref{fig:data-gen}) to construct a dataset of 5K+ Social Stories by prompting advanced LLMs (e.g., GPT-4o) through a breadth-first, hierarchical synthesis from a seed set of manually-written stories, followed by rigorous human filtering. The \textsc{StarSow} is similar to writing a book or sowing a star-tree. Concretely, we first generate chapters to cover various story themes (roots), then create story titles (branches) for each theme as much as possible, and last generate story content (star fruits) cautiously that meets specific constraints given the title and the theme. The analysis shows that our dataset is more diverse, flexible, and effective, making it acceptable to human evaluators. (2) Additionally, to verify the effectiveness of generated Social Stories and ensure their adherence to the strict constraints above, we propose the \textbf{quality assessment criteria} which is employed in both human and GPT evaluations. (3) Finally, we train smaller language models on our crafted dataset, considering that powerful closed-source LLMs require very complex instructions and expensive API fees. Empirical results show that our dataset significantly improves LLMs performance on SS-GEN, at lower costs and with simpler instructions.

In summary, to the best of our knowledge, we are the first to propose SS-GEN task and framework, which aims to make Social Stories automated, affordable, more accessible and of high quality for their use as a social-skill intervention tool within the autism community.
Our framework includes:
\begin{itemize}
    \item Dataset construction by \textsc{StarSow}: A breadth-first, hierarchical, rapid-scaling, and constraint-driven synthetic strategy focusing on qualified Social Stories.
    \item Quality assessment criteria: Personalized criteria for evaluating the adherence of generated Social Stories to the strict constraints.
    \item Empirical results: Experiments by training and testing smaller and cost-effective language models, which show that our dataset significantly improve the LLMs performance on SS-GEN.
\end{itemize}
We hope this work supports autistic youth's social learning needs and inspires future research on special groups.

\section{Social Story Generation}

\begin{figure}[t]
  \includegraphics[width=\linewidth]{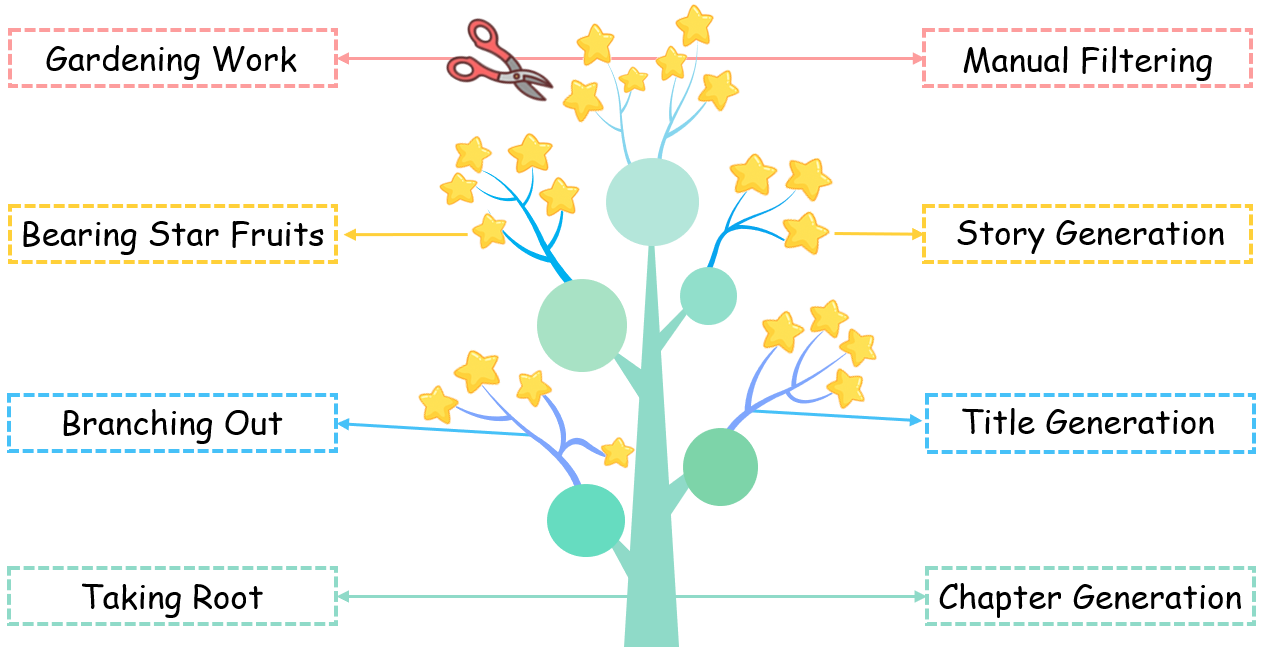}
  \caption{The overview of \textsc{StarSow}, which expands Social Stories in a diverse and hierarchical manner within constraints,  similar to planning a star tree.}
  \label{fig:data-gen}
\end{figure}

In this section, we introduce the development of Social Story Generation (SS-GEN) framework. We begin by formulating the general notation. Next, we explain the dataset synthesis by our refined constraint-driven strategy named \textsc{StarSow} (Figure~\ref{fig:data-gen}), which hierarchically generate varied Social Stories using LLMs (eg., GPT-4o) from a seed set of manually-written stories. Additionally, we propose the quality assessment criteria to verify the adherence of generated Social Stories to the strict constraints.

\subsection{Overview}
A Social Story includes a clear structure: a title that establishes the intervention goal of the story, an introduction that explains the story theme and setting, a main body that depicts social situations and appropriate responses, and a conclusion that summarizes the core idea presented. The last three parts constitute the Social Story content $y$, which is written based on the Social Story title $x$. For SS-GEN, we define the Social Story pair as $(x, y)$ and specify that the language model $M$ is expected to produce the story content $y$ given the story title $x$.


\begin{table*}[t]
\centering
\resizebox{\textwidth}{!}{
\begin{tabular}{|>{\centering\arraybackslash}p{2cm}|p{6cm}|p{12cm}|}
\hline
\textbf{Category} & \textbf{Specific Questions} &  \textbf{Explanation} \\ \hline

\multirow{12}*{\shortstack{\textbf{Structural} \\ \textbf{Clarity} \\ \textbf{($\mathbb{SC.}$)} \\ $1\sim5$ scale} } & \textbf{Q1}: Does the Social Story have a clear structure?& A clear structure refers to that there is a straightforward title establishing the intervention goal, an introduction explaining the story theme and setting, a main body depicting the specific situation and appropriate responses, and a conclusion summarizing the core information presented. \\ \cline{2-3}
& \textbf{Q2:} Do the introduction and the main body show correlation with each other? & The introduction presents the main point, attracts the audience's attention, and guides them to the main body. The main body develops the core idea mentioned in the introduction in detail. \\ \cline{2-3}
&  \textbf{Q3:} Do the main body and the conclusion show a correlation with each other? & The conclusion should highlight the core information from the main body, reinforcing the primary messages and insights discussed. \\ \cline{2-3}
& \textbf{Q4:} Do the conclusion and the introduction show a correlation with each other? & The conclusion and the introduction share the same main point, with the conclusion serving as a callback to the introduction. \\ \hline 

\multirow{6}*{\shortstack{\textbf{Descriptive} \\ \textbf{Orientation} \\ \textbf{($\mathbb{DO.}$)} \\ Yes / No}}  & \textbf{Q1:} Does the Social Story describe more than direct? \, \, \, \, $\frac{\# Descriptive Sentences}{\# Coaching Sentences} \geq 2$ & The number of descriptive sentences should exceed twice the number of coaching sentences in a Social Story according to GR-Eight Formula\citep{gray2010new}. A descriptive sentence accurately describes the interaction, an event, or an explanation of the rationale that underlies what and why people think, say, or do, including perspective and affirmative sentences. Coaching sentences gently and safely guide the behavior. \\ \hline

\multirow{15}*{\shortstack{\textbf{Situational}\\ \textbf{Safety}\\\textbf{($\mathbb{SS.}$)} \\ Yes / No}} & \textbf{Q1:} Does the Social Story use the appropriate person perspective? & \textbf{A)}: The Social Story should never use the second-person perspective to prevent being overly direct. \textbf{B)}: When describing negative behaviors, the Social Story should never employ the first-person perspective to safeguard the dignity and esteem of the audience. \\  \cline{2-3}
& \textbf{Q2:} Does the Social Story consistently convey a positive and patient tone? & The Social Story should always describe situations or guide behaviors in a positive manner, especially when depicting typical or expected behaviors in specific situations. \\ \cline{2-3}
& \textbf{Q3:} Does the Social Story express accurately? & The Social Story should use language that is as clear and unambiguous as possible because ASD children typically interpret things literally rather than inferring implicit meanings that require social insight. \\ \cline{2-3}
& \textbf{Q4:} Does the Social Story use exact vocabulary? & The Social Story should choose the most comfortable and accurate vocabulary for the audience. Firstly, use positive verbs while also being mindful of the varying implications of verbs. Avoid using terms that are likely to evoke strong emotional feelings such as ``shouldn't", ``must", ``supposed to" and so on. \\ \hline
\end{tabular}
}
\caption{The quality assessment criteria table outlines a checklist of questions for Structural Clarity ($\mathbb{SC.}$), Descriptive Orientation ($\mathbb{DO.}$), and Situational Safety ($\mathbb{SS.}$), each accompanied by a detailed explanation.}
\label{tab:rating-scale}
\end{table*}

\subsection{Dataset Construction}
\label{sub:starsow}
To expand qualified Social Stories with broad coverage, we develop a constraint-driven strategy \textsc{StarSow}, which hierarchically expands and synthesizes data as if planting a star tree. The dataset we want to generate contains a set of root nodes $\{(C_t, E_t)\}$ respectively representing the chapters and their explanations that define Social Story themes. Each root node $(c_t, e_t)$ further branches out into multiple Social Story pairs $\{x_{t,j}, y_{t,j}\}_{j=1}^{n_t},n_t>1$, where $x_{t,j}$ and $y_{t,j}$ mean the title and the content of the $j^{th}$ Social Story in the chapter node $(c_t, e_t)$.

The process for \textsc{StarSow} strategy consists of four steps: 1) Taking root: explain and expand the chapters, 2) Branching Out: generate diverse Social Story titles from the root layer, 3) Bearing Star Fruits: complete the Social Story content given the title, 4) Gardening Work: filter out chapters, titles, and stories which are invalid and redundant.

\noindent\textbf{Taking Root.}
\textsc{StarSow} begins by generating new chapter root nodes using an expert-written seed set as a reference \citep{gray2010new}. This seed set contains 14 chapters and 179 meticulously selected Social Story pairs in total. To ensure the soundness of new chapter themes, 
we apply an explain-then-generate approach to better leverage the emergent abilities of the LLM. Firstly, we prompt the LLM to obtain broad explanations of the seed chapters, which form the foundational seed set of root nodes. Secondly, we sample 8 ``chapter-explanation" pairs as in-context examples to expand each new root node. Of the 8 examples, 4 are from the seed set, and 4 are from the model-generated brand-new nodes in previous steps to promote diversity.   

\noindent\textbf{Branching Out.}
As a certain number of new root nodes is reached, we branch out the corresponding Social Story titles for each node to cover broader intervention goals. This requires the LLM to understand the diverse range of issues that autistic youth may encounter within each specific chapter's theme. We found that the LLM can come up with diverse story titles to a large extent when prompted with 8 ``chapter-explanation-titles" in-context examples which randomly sampled from the seed set.

\noindent\textbf{Bearing Star Fruits.}
As the chapter root nodes grow, and the story titles flourish, we ultimately obtain qualified story fruits by carefully controlling the cultivation conditions. To positively improve social understanding and behavioral outcomes for autistic youth, each Social Story must strictly adhere to our summarized specific constraints for SS-GEN, according to the 10.2 Tutorial \citep{gray2010new} which provides detailed guidance on writing and applying Social Stories. These constraints include structural clarity, descriptive orientation, and situational safety.
Additionally, we sampled 4 ``chapter-explanation-title-story" from the seed set as qualified in-context examples. Finally, we designed a constraint-driven prompt to guide the LLM in completing the Social Story given the root nodes with ``chapter-explanation" pair and the corresponding Social Story ``title". 

\noindent\textbf{Gardening Work.}
To encourage diversity, a new chapter is added to the Social Story dataset only when its ROUGE-L similarity with any existing chapter is less than 0.7. The same applies to a new Social Story title. When generating the Social Story content based on the title goal, we filter out invalid outputs that do not meet the strict constraints of the Social Stories (e.g., using the narrative ``you", or negative words that usually cause anxious or resistant feelings). After manual screening and refinement, only 74.3\% of the data was deemed valid to construct the final dataset.

\subsection{Quality Assessment Criteria}
\label{sec:quality-criteria} 
To evaluate the validity of Social Stories and their adherence to strict constraints, we design the quality assessment criteria in Table~\ref{tab:rating-scale} based on the constraints of structural clarity, descriptive orientation, and situational safety. Each constraint includes corresponding checklist questions with detailed explanations provided afterward. We define structural clarity as a soft constraint, scored on a 1 to 5 point scale, with 5 being the best. The remaining constraints are considered hard constraints and assessed with ``Yes / No". 

Additionally, we have developed a website for manual assessment of Social Stories and preference ranking based on the above rating scale, which will be utilized in assessing the quality of generated data in subsequent sections.

\begin{table}[tbp]
  \centering
  
  \small
    \begin{tabular}{lc}
    \toprule
    \multicolumn{2}{p{23em}}{Statistic} \\
    \midrule
    \# of chapters  & 57 \\
    \# of titles in each chapter & $\geq$70 \\
    \# of Social Story Pairs in total & 5085 \\
    ave.chapter length (in words) &  2.46 \\
    ave.title length (in words) & 5.28 \\
    ave.story content length (in words) & 281.65 \\
    \bottomrule
    \end{tabular}%
  
  \caption{Statistics of the constructed Social Story dataset by applying \textsc{StarSow} to GPT-4.}
  \label{tab:statistics}%
\end{table}%

\begin{figure}[t]
    \centering
  \includegraphics[width=0.8\columnwidth]{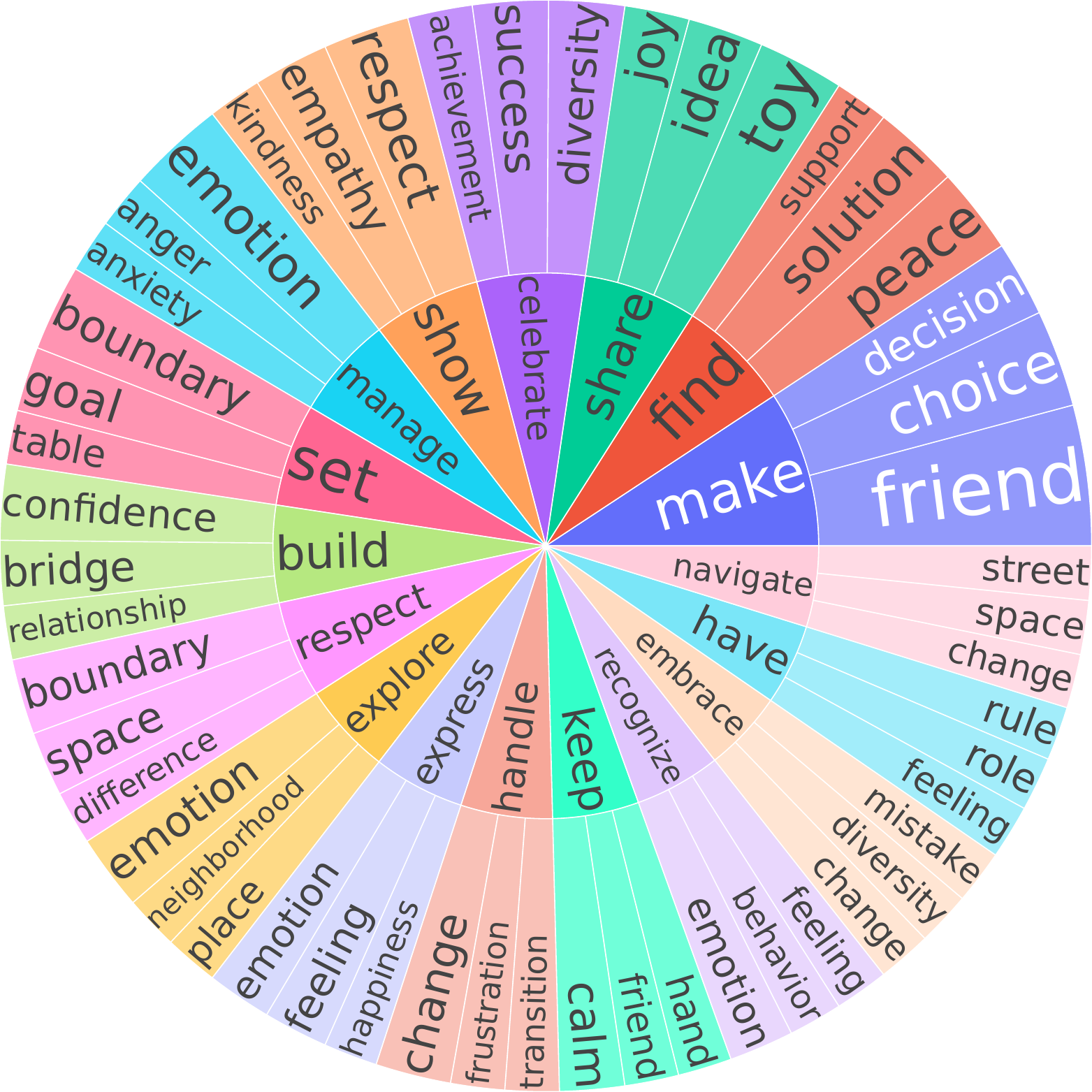}
  \caption{The top 20 most common root verbs (inner circle) and their top 3 direct noun objects (outer circle) in the generated Social Story titles.}
  \label{fig:verb-noun}
\end{figure}

\section{Assessment of Social Story}
\label{sec:dataset-quality}
This section provides an overview of the Social Story dataset created using \textsc{StarSow}. We select the engine \texttt{GPT-4o}\footnote{OpenAI API: \url{https://openai.com/api/}. Query parameters are detailed in the supplementary.} as the data synthesizer based on the quality analysis below.

\subsection{Statistics}
Table~\ref{tab:statistics} describes the basic statistics of the generated dataset for SS-GEN. We generated a total of 57 chapter themes and more than 5K Social Stories, with at least 70 in each chapter after filtering. We then divide the data into training, validation, and testing sets with a ratio of 8:1:1.

\begin{figure}[t]
    \centering
    \begin{subfigure}[b]{0.45\textwidth}
        \centering
        \includegraphics[width=0.9\textwidth]{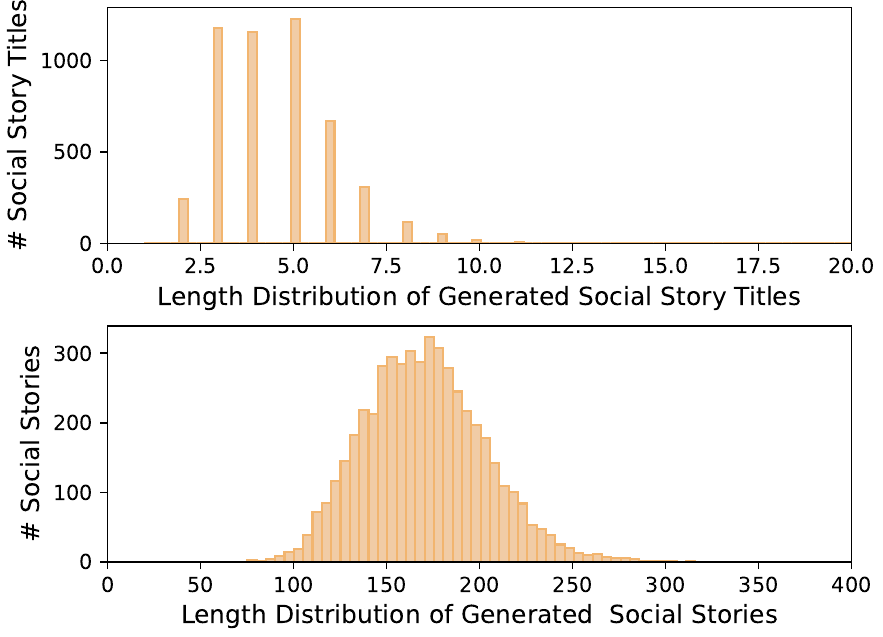}
        \caption{Length distribution of the generated Social Story pairs.}
        \label{fig:length-statistic}
    \end{subfigure}
    \hfill
    \begin{subfigure}[b]{0.45\textwidth}
        \centering
        \includegraphics[width=0.9\textwidth]{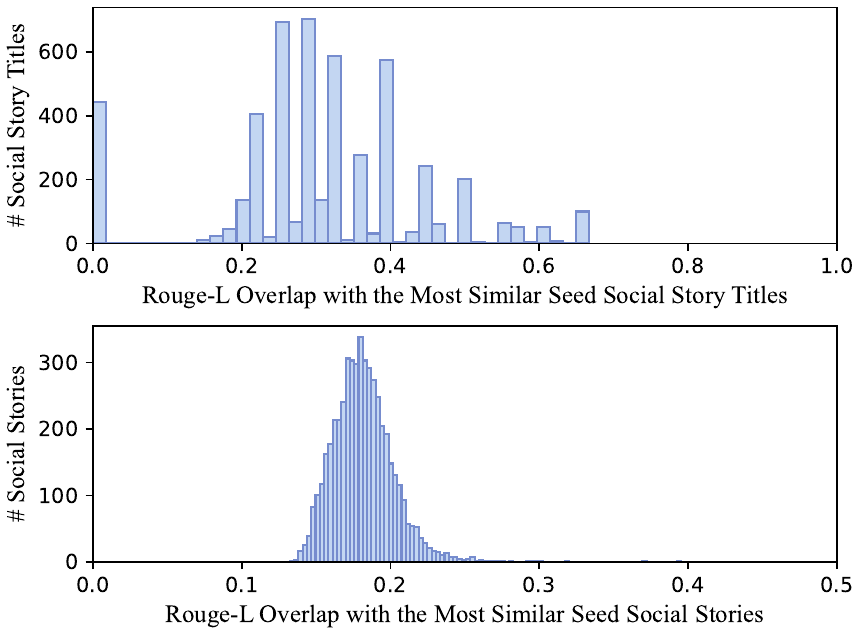}
        \caption{Distribution of the ROUGE-L scores between the generated Social Stories and their most similar seed stories.}
        \label{fig:rougeL-statistic}
    \end{subfigure}
    \caption{Overview of generated data statistics.}
    \label{fig:combined}
\end{figure}
\subsection{Diversity}
We explore the distinctions between the synthesized dataset and the manually written seed set that prompted the generation. For each generated Social Story pair (title, story content), we calculate its highest ROUGE-L overlap with the corresponding data in the seed set. Figure~\ref{fig:rougeL-statistic} shows the distribution of these ROUGE-L scores, indicating that a significant number of new titles and stories were generated with minimal overlap with the seeds. Additionally, we also examine diversity from the length perspective, with the distribution shown in Figure~\ref{fig:length-statistic}.

To further analyze what types of intervention goals are generated and how diverse they are, we identify the verb-noun structure in the generated Social Story titles. Through the Berkeley Neural Parser\footnote{https://parser.kitaev.io/} \cite{kitaev2018constituency,kitaev2019multilingual}, we parse the titles and then extract the verb closest to the root along with its first direct noun object. Figure~\ref{fig:verb-noun} shows the top 20 most common root verbs and their top 3 direct noun objects, which can represent common topics within the dataset. The figure shows these intervention goals exhibit a wide range of intents and textual formats.

\subsection{Quality}
\label{sec:quality}

\begin{table}[tbp]
  \centering
  
  \fontsize{8pt}{9pt}\selectfont 
    \resizebox{0.95\columnwidth}{!}{
    \begin{tabular}{>{\centering\arraybackslash}c|m{18em}cc}
    \toprule
    & \centering \small\textbf{Quality Assessment Criteria \& Human Preference} & \multicolumn{1}{m{2.4em}}{GPT3} & \multicolumn{1}{m{2.4em}}{GPT4} \\ \hline
    \multirow{7}*{$\mathbb{SC.}$} &Does the Social Story have a clear structure? &  82.7\%    & 83.0\%  \\ \cline{2-4}
    &Does the introduction and the main body show a correlation with each other? &  86.2\%   & 86.8\% \\ \cline{2-4}
    &Does the main body and the conclusion show a correlation with each other? &  82.5\%   & 83.7\% \\ \cline{2-4}
    &Does the conclusion and the introduction show a correlation with each other? &  84.1\%   & 82.9\% \\ \hline
    
    $\mathbb{DO.}$ &Does the Social Story describe > direct? &  56.0\%    & 59.0\% \\ \hline
    
    \multirow{5}*{$\mathbb{SS.}$} &Does the Social Story use the appropriate person perspective? (A) &  90.0\%  & 94.0\% \\ \cline{2-4}
    &Does the Social Story use the appropriate person perspective? (B) &  91.0\%  & 93.0\% \\  \cline{2-4}
    &Does the Social Story consistently convey a positive and patient tone? & 97.0\%   & 98.0\% \\ \cline{2-4}
    &Does the Social Story expresses accurately? &  91.0\%  & 90.0\% \\ \cline{2-4}
    &Does the Social Story use exact vocabulary? &  95.0\%  & 92.0\% \\ 
    \hline
    
    $\mathbb{HP.}$ & How do you prefer the Social Story? &  38.0\%  & 62.0\% \\ 
    \bottomrule
    \end{tabular}%
    }
  \caption{Data quality review of Social Stories generated from GPT-3 and GPT-4 given titles in the seed set. $\mathbb{SC.}$: Structural Clarity, $\mathbb{DO.}$: Descriptive Orientation, $\mathbb{SS.}$: Situational Safety, $\mathbb{HP.}$: Human Preference.}
  \label{tab: rating}%
\end{table}%

We have shown the quantity and diversity of the generated data before, but the quality remains uncertain, which is crucial for its effectiveness in social interventions for autistic youth. 
To assess the quality of Social Stories generated by the GPT series, we randomly sample 100 story pairs from the original seed set, using their titles to prompt \texttt{gpt-3.5-turbo}, \texttt{gpt-4o} respectively and regenerate stories as like \textbf{Bearing Star Fruits}. The two sets of regenerated stories are then compared with the story references from the sampled pairs. Specifically, we carefully check the two sets based on checklist questions in Table~\ref{tab:rating-scale}, and then rank the two sets based on the preference.


Table~\ref{tab: rating} presents the assessment results focusing on structural clarity ($\mathbb{SC.}$), descriptive orientation ($\mathbb{DO.}$), situational safety ($\mathbb{SS.}$) and the human preference ($\mathbb{HP.}$). The $\mathbb{SC.}$ depicts the percentage of the average score relative to the full score. The $\mathbb{DO.}$ and $\mathbb{SS.}$ show the percentage of meeting the corresponding criteria. The $\mathbb{HP.}$ represents the percentage of human preference between the two sets, summing to 1.

The findings reveal that both engines score over 80\% on $\mathbb{SC.}$ and $\mathbb{SS.}$ and can perform well on SS-GEN, indicating GPT series can construct logically structured Social Stories and ensure the narrative's safety and appropriateness, despite subtle differences in narrative flow and language nuances. Although the two sets perform relatively lower on $\mathbb{DO.}$ with scores between 50\% to 60\%, we found that those stories are indeed not prescriptive or disrespectful, and they can be easily improved with simple modifications. Notably, \texttt{gpt-4o} consistently outperforms \texttt{gpt-3.5-turbo} in most categories and is preferred more (62\%). Therefore we finally choose \texttt{gpt-4o} to construct the whole dataset. 

\section{Experiments}
\label{sec:experimental-setup}
In this section, we conduct extensive experiments\footnote{The code, prompt, data and technical appendix are available at \url{https://github.com/MIMIFY/SS-GEN}} to evaluate the performance of smaller, cost-effective language models which are trained and tested using our curated dataset for SS-GEN, considering that powerful closed-source LLMs require very complex instructions and expensive API fees.


\subsection{Experimental Settings}
\noindent\textbf{Model selection:}
To conduct thorough experiments within the controlled resource, we selected a series of mainstream language models with 2B, 7B and 8B parameters. Specifically, we choose \textbf{Mistral} (2B, 7B) \citep{jiang2023mistral}, \textbf{Gemma} (2B, 7B) \citep{team2024gemma}, \textbf{LLama3-8B}\footnote{https://github.com/meta-llama/llama3} along with each variant of instruction tuning (Instruct). 

The experiments include two types. Firstly, we test the performance of popular language models on SS-GEN by directly prompting the original models to generate the Social Story content based on the title (Zero-Shot). Next, we fine tune these models with the proposed dataset and then compare the generated Social Stories (SFT).  

We utilize the Parameter-Efficient Fine-Tuning (PEFT) strategy, integrated with Low-Rank Adaption (LoRA) using the LLaMA Factory \citep{zheng2024llamafactory}, to test and fine-tune these models on four NVIDIA GeForce RTX 4090 GPUs.

\noindent\textbf{Dataset and prompt:} 
We utilize the Social Story dataset constructed through \textsc{StarSow} for SS-GEN. The dataset is split with 80\%, 10\%, 10\% respectively for training, validating and testing. 

Besides, we utilize the same precise ``title-to-story" prompt as illustrated in Figure~\ref{prompt:title-to-story} for both training and testing. This simple prompt is designed to enhance the model's capacity to construct a Social Story from the provided title (the intervention goal of the story). 


\definecolor{mycolor}{HTML}{e7e2e8} 
\colorlet{mycolor}{mycolor!10} 
\begin{figure}
  \centering
  \resizebox{0.8\columnwidth}{!}{
  \begin{tcolorbox}[
      boxsep=0pt,
      interior style=empty,
      left=5pt, right=5pt,
      before skip=0pt, 
      after skip=0pt,
      boxrule=0.9pt, 
      halign = flush center, 
      nobeforeafter, 
      colback=mycolor 
    ]
    \small
    \textbf{\texttt{Title-to-Story for SS-GEN}}
    
    \tcblower 
    \small
    Develop a concise, clear, straightforward, positive and supportive Social Story titled ``\{\texttt{title}\}" for children and teens with autism, 200-300 words, that promotes their social understanding and boosts their participation in daily activities, fostering independence and confidence.\newline{}
    Title: \texttt{ \{title\}}
  \end{tcolorbox}}
  \caption{The prompt is designed to enhance the language model's ability to complete the Social Story content given the Social Story title.}
  \label{prompt:title-to-story}
\end{figure}

\begin{figure*}[tbp]
\centering
  \includegraphics[width=0.98\textwidth]{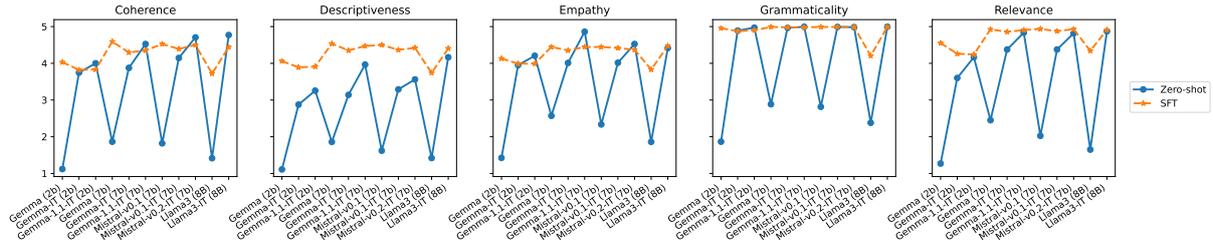}
  \caption{GPT-4 evaluation results of various models on five dimensions on Zero-Shot (blue) and SFT (orange) scenarios for  SS-GEN: Coherence ($\mathcal{CH.}$), Descriptiveness ($\mathcal{DC.}$), Empathy ($\mathcal{EM.}$), Grammaticality ($\mathcal{GA.}$), and Relevance ($\mathcal{RE.}$).}
  \label{fig:gpt4-evaluation}
\end{figure*}

\begin{table*}[tbp]
  \centering
  
  \resizebox{0.98\textwidth}{!}{
    \begin{tabular}{p{3.65cm}|ccccc|ccccc}
    \toprule
    \multicolumn{1}{c|}{\multirow{2}{*}{ Model}} & \multicolumn{5}{c|}{Zero-Shot (\%)} & \multicolumn{5}{c}{SFT (\%)} \\
    \cmidrule(lr){2-6} \cmidrule(lr){7-11}
          & BLEU-4 & ROUGE-1 & ROUGE-2 & ROUGE-L & BertScore  & BLEU-4 & ROUGE-1 & ROUGE-2 & ROUGE-L & BertScore \\
    \midrule
    Gemma (2B) & 8.86  & 21.81 & 4.63  & 7.22  & 79.0 & $\textbf{51.18}_{+42.32\%}^{\dagger}$ & $\textbf{47.58}_{+25.77\%}^{\dagger}$ & $\textbf{20.13}_{+15.50\%}^{\dagger}$ & $\textbf{27.29}_{+20.07\%}^{\dagger}$ & $ 88.8_{+9.8\%}^{\dagger}$ \\
    Gemma-Instruct (2B) & 31.68 & 30.44 & 6.49  & 14.17 & 82.7 &  $50.35_{+18.67\%}^{\dagger}$ & $46.53_{+16.09\%}^{\dagger}$ & $19.36_{+12.87\%}^{\dagger}$ & $26.93_{+12.76\%}^{\dagger}$ & $88.8_{+6.1\%}^{\dagger}$ \\
    Gemma-1.1-Instruct (2B) & \textbf{42.43} & \textbf{34.98} & \textbf{8.82}  & \textbf{17.22} & \textbf{84.9} & $51.02_{+8.590\%}^{\dagger}$ & $46.^{\dagger}70_{+11.72\%}^{\dagger}$  & $19.53_{+1^{\dagger}0.71\%}^{\dagger}$ & $27.15_{+9.930\%}^{\dagger}$ & $\textbf{88.9}_{+4.0\%}^{\dagger}$\\
    \midrule
    Gemma (7B) & 19.24 & 32.12 & 9.57  & 12.48 & 81.7 & $54.74_{+35.50\%}^{\dagger}$ & $\textbf{52.12}_{+20.00\%}^{\dagger}$ & $24.74_{+15.17\%}^{\dagger}$ & $31.04_{+18.56\%}^{\dagger}$ & $\textbf{89.9}_{+8.2\%}^{\dagger}$ \\
    Gemma-Instruct (7B) & 39.80 & 32.56 & 8.09  & 16.33 & 84.1 &  $53.82_{+14.02\%}^{\dagger}$ & $50.80_{+18.24\%}^{\dagger}$  & $23.34_{+15.25\%}^{\dagger}$ & $30.19_{+13.86\%}^{\dagger}$ & $89.7_{+5.6\%}^{\dagger}$\\
    Gemma-1.1-Instruct (7B) & \textbf{45.64} & 37.65 & 10.00  & 17.87 & 84.8 &  $54.04_{+8.400\%}^{\dagger}$ & $51.04_{+13.39\%}^{\dagger}$ & $23.71_{+13.71\%}^{\dagger}$ & $30.40_{+12.53\%}^{\dagger}$ & $89.8_{+5.0\%}^{\dagger}$\\
    Mistral-v0.1 (7B) & 19.76 & 25.78 & 5.95  & 12.47 & 81.6 &  $\textbf{54.84}_{+35.08\%}^{\dagger}$ & $51.70_{+25.92\%}^{\dagger}$  & $24.07_{+18.12\%}^{\dagger}$ & $30.58_{+18.11\%}^{\dagger}$ & $89.8_{+8.6\%}^{\dagger}$\\
    Mistral-v0.1-Instruct (7B) & 40.14 & 38.23 & 11.85 & \textbf{20.79} &\textbf{86.5} &  $54.17_{+14.03\%}^{\dagger}$ & $51.17_{+12.94\%}^{\dagger}$ & $23.79_{+11.94\%}^{\dagger}$ & $30.54_{+9.750\%}^{\dagger}$ & $89.7_{+3.2\%}^{\dagger}$ \\
    Mistral-v0.2-Instruct (7B) & 37.44 & 37.70  & 11.51 & 19.74 & 86.3 &  $54.30_{+16.86\%}^{\dagger}$  & $52.05_{+14.35\%}^{\dagger}$ & $\textbf{24.75}_{+13.24\%}^{\dagger}$ & $\textbf{31.18}_{+11.44\%}^{\dagger}$ & $\textbf{89.9}_{+3.6\%}^{\dagger}$ \\
    LLama3 (8B) & 9.49  & 23.46 & 5.82  & 8.15 & 79.3 &   $16.41_{+6.920\%}$ & $47.66_{+24.20\%}^{\dagger}$ & $22.20_{+16.38\%}^{\dagger}$  & $9.69_{+1.540\%}$ & $81.2_{+1.9\%}$\\
    LLama3-Instruct (8B) & 39.04 & \textbf{41.19} & \textbf{12.68} & 19.37 & 85.5 &  $54.56_{+15.52\%}^{\dagger}$ & $51.26_{+10.07\%}^{\dagger}$ & $23.64_{+10.96\%}^{\dagger}$ & $30.30_{+10.93\%}^{\dagger}$ & $89.7_{+4.2\%}^{\dagger}$\\
    \bottomrule
  \end{tabular}
  }
  \caption{Average Performance of three inferences on traditional metrics in Zero-Shot and SFT scenarios for SS-GEN. The subscripts on the bottom right indicate the performance change of the same model after fine-tuning with the dataset. The symbol (${\dagger}$) indicates a significant difference between pre- and post-finetuning results ($p < 0.01$) in a paired t-test at $\alpha = 0.05$.} 
  \label{tab:zero-sft}
\end{table*}

\subsection{Evaluation Methods}
To comprehensively and reliably evaluate the generated Social Stories, we leverage both objective measures from traditional metrics and qualitative evaluations based on the proposed quality assessment criteria.

\noindent\textbf{Traditional Metrics: }
We use a variety of evaluation metrics, including word-based BLEU-4, ROUGE-1, ROUGE-2, ROUGE-L scores, and embedding-based BertScore, to ensure objective and quantifiable assessments. Higher values in these metrics indicate better model performance and greater N-gram similarity between the generated text and the reference text, showing that the generated data more accurately reflects the patterns found in the reference.

\noindent\textbf{Quality Assessments: }
We incorporate the proposed quality assessment criteria for Social Stories into both GPT-4 and human evaluations. 

For GPT-4 evaluation, we apply the Coherence ($\mathcal{CH.}$), Descriptiveness ($\mathcal{DC.}$), Empathy ($\mathcal{EM.}$) scores to separately evaluate whether the Social Story meets the checklist requirements for structural clarity, descriptive orientation, situational safety as outlined in the established criteria for quality assessment. Additionally, for the text itself, we use the Grammaticality ($\mathcal{GA.}$) score to determine grammatical correctness, fluency, and adherence to language standards. We also use the Relevance ($\mathcal{RE.}$) score to assess the consistency and alignment between the generated story content and the given title prompt. All scores range from 1 to 5, with 5 representing the highest level of quality, clarity, and adherence to the evaluation standards.

For human evaluation, we invite three psychology students who received comprehensive training on the quality assessment criteria guidelines before their participation in the evaluation. Additionally, we develop a user-friendly website shown in the supplementary to standardize expert evaluations, improve efficiency, and enhance the horizontal comparison of text generated by different models. Based on the criteria, experts use this website to score Q1, Q2, Q3, Q4 of the structural clarity ($\mathbb{SC.}$) on a scale of 1 to 5, and to provide ``Yes/No" for Q1 of the descriptive orientation ($\mathbb{DO.}$) and Q1(A), Q1(B), Q2, Q3, Q4 of the situational safety ($\mathbb{SS.}$). Finally, we report the average score of all $\mathbb{SC.}$ sub-items, as well as the proportion of qualified $\mathbb{SS.}$ and $\mathbb{DO.}$, where the ``qualified" means all sub-items are rated as ``Yes".

\subsection{Results and Analysis}
\label{sec:experimental-results}

We evaluate mainstream models of various scales for generating Social Stories in Zero-Shot and SFT scenarios. Table~\ref{tab:zero-sft} presents the average results of three inferences on the test set using traditional metrics. Figure~\ref{fig:gpt4-evaluation} shows GPT-4 evaluations of 200 randomly selected stories from the test set, focusing on Coherence ($\mathcal{CH.}$), Descriptiveness ($\mathcal{DC.}$), Empathy ($\mathcal{EM.}$), Grammaticality ($\mathcal{GA.}$) and Relevance ($\mathcal{RE.}$). 
In Table~\ref{tab:human-evaluation}, we select the top 2B and 7B models on both traditional and GPT-4 evaluations for further expert assessment, along with 50 randomly selected stories from the test set for the expert assessment. We present the analysis of the case study in the technical appendix.

\noindent\textbf{Performance improvement through SFT:} 
We significantly improve LLMs on SS-GEN through fine-tuning with our proposed Social Story dataset, which can be shown by the positive changes in the subscripts on the bottom right of SFT scenarios in Table~\ref{tab:zero-sft}. For example, SFT scores in Table~\ref{tab:zero-sft} are markedly higher than Zero-Shot scores every row. The fine-tuned Gemma (2B) improves its performance by 42.32\%\textuparrow (BLEU-4), 25.77\%\textuparrow (ROUGE-1), 15.50\%\textuparrow (ROUGE-2), 20.07\%\textuparrow (ROUGE-L) and 9.8\%\textuparrow (BertScore). Multiple inferences (3 times) were conducted for each model before and after fine-tuning, followed by a t-test ($\alpha=0.05$), with ${\dagger}$ in Table~\ref{tab:zero-sft} indicating a significant difference between pre- and post-fine-tuning results ($p < 0.01$).

As for GPT-4 quality evaluation, Figure~\ref{fig:gpt4-evaluation} shows that the yellow line representing SFT models  is consistently higher than the blue line representing Zero-Shot models. Moreover, we select the overall best-performing 2B/7B models on both traditional evaluations and GPT-4 evaluations for further expert assessment, which are fine-tuned Gemma (2B/7B), zero-shot Gemma-1.1-Instruct (2B/7B) for human evaluation, shown in Table~\ref{tab:human-evaluation}. Results from human evaluation indicates that the stories generated by Zero-Shot models have a much lower situational safety ($\mathbb{SS.}$) compliance rate compared to those generated by SFT models.
Besides, Table~\ref{tab:Pearson Correlation} shows the correlation between ratings provided by GPT evaluators ($\mathcal{CH.}$, $\mathcal{DC.}$, $\mathcal{EM.}$) and human evaluators ($\mathbb{SC.}$, $\mathbb{DO.}$, $\mathbb{SS.}$) on the same set of data as Table~\ref{tab:human-evaluation}, indicating a positive correlation trend between the GPT evaluation and human evaluation.

\noindent\textbf{Instruct Models on Zero-Shot:} In Zero-shot scenarios, Instruct models consistently demonstrate superior performance compared to their original pre-trained base counterparts. This is evident in both traditional metrics and GPT-4 evaluation. For instance, Figure~\ref{fig:gpt4-evaluation} shows that all peaks of the blue lines correspond to Instruct models, while all troughs indicate base models. Besides, Table~\ref{tab:zero-sft} demonstrates that under Zero-Shot settings, Gemma-1.1-Instruct (2B) improves performance by 33.57\%\textuparrow (BLEU-4), 13.17\%\textuparrow (ROUGE-1), 4.19\%\textuparrow (ROUGE-2), and 10\%\textuparrow (ROUGE-L) compared to Gemma (2B).

\begin{table}[tbp]
  \centering
  
  \resizebox{0.95\columnwidth}{!}{
    \begin{tabular}{c|c|rrr}
    \toprule
    \multicolumn{1}{c|}{\multirow{2}[4]{*}{Scenarios}} & \multicolumn{1}{c|}{\multirow{2}[4]{*}{Model}} & \multicolumn{3}{c}{Quality Assessment}  \\
\cmidrule{3-5}          &       & \multicolumn{1}{c}{$\mathbb{SC.}$} & \multicolumn{1}{c}{$\mathbb{DO.}$} & \multicolumn{1}{c}{$\mathbb{SS.}$} \\
    \midrule
    \multirow{2}[2]{*}{SFT} & Gemma (7B) & 4.90  & 86.7\% & 83.3\%  \\
          & Gemma (2B) & 4.83  & 83.3\% & 80.0\%  \\
    \midrule
    \multirow{2}[2]{*}{Zero-Shot} & Gemma-1.1-Instruct (7B) & 3.89  & 93.3\%  & 10.0\%  \\
          & Gemma-1.1-Instruct (2B) & 3.28 & 90.0\%  & 6.00\%  \\
    \bottomrule
    \end{tabular}%
    }
    \caption{Results of the human evaluation on quality assessment and preference ranking for the best Gemma models (2B, 7B) on both Zero-Shot and SFT scenarios.}
  \label{tab:human-evaluation}%
\end{table}%

\begin{table}[t]
\centering

\resizebox{0.95\columnwidth}{!}{
\begin{tabular}{c|cccc} 
\toprule
\textbf{Scenario} & \textbf{Model} & $\mathbb{SC.}$ / $\mathcal{CH.}$ & $\mathbb{DO.}$ / $\mathcal{DC.}$ & $\mathbb{SS.}$ / $\mathcal{EM.}$ \\
\midrule
\multirow{2}{*}{SFT} & Gemma (7B) & 0.52 & 0.43 & 0.47 \\
 & Gemma (2B) & 0.49 & 0.46 & 0.34 \\
\midrule
\multirow{2}{*}{Zero-Shot} & Gemma-1.1-Instruct (7B) & 0.54 & 0.44 & 0.41 \\
 & Gemma-1.1-Instruct (2B) & 0.51 & 0.48 & 0.36 \\
\bottomrule
\end{tabular}
}
\caption{Pearson correlation coefficients between GPT and human Ratings, showing a positive correlation trend.}
\label{tab:Pearson Correlation}
\end{table}

\noindent\textbf{Base Models on SFT:} In SFT scenarios, base models demonstrate greater potential for adapting to domain-specific data. For example, fine-tuned base models outperform fine-tuned Instruct models, as seen in both GPT evaluations and traditional metrics. This trend holds true for the models of Gemma and Mistral series, but not for the LLama3 series. It is speculated that the LLama3 base model, pre-trained with 15T data, may need a larger and more targeted fine-tuning dataset to effectively leverage its pre-learned complex patterns and features, thus reaching optimal performance on specific tasks like SS-GEN.

\section{Related Work}
\noindent\textbf{LLMs for Story Generation.} The advent of LLMs has significantly advanced automatic story generation, enabling tailored narratives with minimal manual intervention to mimic human-written stories. Researchers have utilized storytelling to enhance various fields, including creating playable 2D story-games \citep{nasir2024word2world}, crafting suspenseful stories \citep{xie2024creating}, generating longer and more creative narratives \citep{yang2023doc}, and conveying legal concepts \citep{jiang2024leveraging}. However, these stories focus on adult audiences, emphasizing narrative flow, entertainment, and emotional engagement. This free-form approach contrasts sharply with the requirements of Social Stories, which need clear structure and accurate description. \citep{valentini2023automatic,bhandari2023trustworthiness} explored generating age-appropriate stories for typical children, assessing the trustworthiness and suitability of LLM-generated content. Nonetheless, these stories do not meet the stringent demands of situational safety in Social Stories, including constraints on narrative perspective, positive tone, vocabulary selection, and precise expression. In this work, we explore the constraint-driven strategy of generating and evaluating Social Stories, focusing on the audience of children with autism.
 
\noindent\textbf{LLMs for Data Augmentation.}
Built upon massive corpora and qualified alignment, the increasingly powerful LLMs \citep{achiam2023gpt} are promising at producing texts of eligible quality, resulting in a surge of research interests in exploring using LLMs to create synthetic data for tasks or to augment existing datasets to boost task performance \citep{patel2024datadreamer}. Such applications include generating tabular data \citep{borisov2022language}, relation triplets \citep{chia2022relationprompt}, sentence pairs \citep{schick2021generating}, open-domain dialogue \citep{zheng2023augesc}, instruction data \citep{peng2023instruction,shao2023synthetic,wang2023self,sun2024principle}, etc.. Enhanced task performance validates the effectiveness of LLM-generated data. In this work, we design a constraint-driven strategy named \textsc{StarSow} to use LLMs for constructing qualified Social Story dataset, aiming to make Social Stories more accessible, affordable, and of high quality for their use as a social-skill intervention tool within the autism community.

\section{Conclusion}
We propose \textbf{SS-GEN}, an innovative Social Story Generation framework that includes: (1) Qualified dataset construction for SS-GEN through a breadth-frist, hierarchical and constraint-driven strategy \textbf{\textsc{StarSow}}. (2) The personalized \textbf{quality assessment criteria} for evaluating the adherence of generated Social Stories to their unique and strict constraints. (3) Extensive experiments by training and testing smaller, cost-effective language models, which show that our dataset significantly improve the LLMs performance on SS-GEN. We hope SS-GEN makes Social Stories more accessible, affordable and of high quality for their use as a social-skill intervention tool within the autism community.

\section{Ethical Statement}
The generation of Social Stories entails a range of ethical considerations and limitations that require careful attention to ensure their responsible application, particularly in sensitive contexts involving neurodiverse children. The primary ethical issues are outlined as follows. (1) Bias and Stereotypes: training data may carry social biases although we have rigorous human review to ensure quality. (2) Caregiver supervision: AI-drafted stories need caregiver's refinement to suit each child's unique experiences, ensuring ethical and emotional suitability. (3) Ethical risks: limited transparency in generation may affect long-term predictability. Expert feedback and heuristic controls are advised to mitigate emotional insensitivity risks.

\section{Acknowledgments}
This work was partly supported by the National Key Research and Development Program of China under Grant 2024YFE0202900; the National Natural Science Foundation of China under Grant (62436001, 62176020); the Joint Foundation of the Ministry of Education for Innovation team (8091B042235); the Fundamental Research Funds for the Central Universities (2019JBZ110) ; and the State Key Laboratory of Rail Traffic Control and Safety (Contract No. RCS2023K006), Beijing Jiaotong University. We sincerely thank our collaborators for their invaluable contributions, including insightful feedback and support in refining the paper and addressing key challenges. Their dedication and expertise greatly enhanced the quality and impact of this work. We thank the anonymous reviewers for carefully reading our paper and their insightful comments and suggestions.

\bibliography{aaai25}

\end{document}